\documentclass[noinfoline]{imsart}
\usepackage{amsthm,amsmath,amsfonts,natbib}
\usepackage{verbatim}
\usepackage{graphicx}
\usepackage{caption}
\usepackage{tabu}
\usepackage{booktabs}
\usepackage{subcaption}
\usepackage{bm}
\usepackage{xcolor}
\usepackage{bbm}

\usepackage{todonotes}
\usepackage{amssymb}
\usepackage{tikz}
\usetikzlibrary{fit,shapes}
\RequirePackage[colorlinks,citecolor=blue,urlcolor=blue]{hyperref}

\startlocaldefs
\newcommand\Tstrut{\rule{0pt}{2.6ex}}  
\newcommand\Bstrut{\rule[-0.9ex]{0pt}{0pt}}
\newcommand{\Gr}{\mathcal{G}}
\newcommand\independent{\protect\mathpalette{\protect\independenT}{\perp}}\def\independenT#1#2{\mathrel{\rlap{$#1#2$}\mkern2mu{#1#2}}}
\endlocaldefs

\newtheorem{example}{Example}
\newtheorem{definition}{Definition}
\newtheorem{lemma}{Lemma}
\newtheorem{proposition}{Proposition}
\newtheorem{theorem}{Theorem}
\DeclareMathOperator*{\argmax}{arg\,max}

\begin{document}

\begin{frontmatter}

\title{Directed expected utility networks}
\runtitle{Directed expected utility networks}


\author{\fnms{Manuele} \snm{Leonelli}\ead[label=e1]{manuele@dme.ufrj.br}}
\address{\printead{e1}}
\affiliation{Universidade Federal do Rio de Janeiro}
\address{Departamento de Estatistica, Universidade Federal do Rio de Janeiro, Rio de Janeiro, Brazil.}
\author{\fnms{Jim Q.} \snm{Smith}\ead[label=e2]{j.q.smith@warwick.ac.uk}}
\address{\printead{e2}}
\affiliation{University of Warwick}
\address{Department of Statistics, University of Warwick, CV47AL Coventry, UK.}
\runauthor{M. Leonelli and J.Q. Smith}

\begin{abstract}
A variety of statistical graphical models have been defined to represent the conditional independences underlying a random vector of interest. Similarly, many different graphs embedding various types of preferential independences, as for example conditional utility independence and generalized additive independence, have more recently started to appear. In this paper we define a new graphical model, called a directed expected utility network, whose edges depict both probabilistic and utility conditional independences. These embed a very flexible class of utility models, much larger than those usually conceived in standard influence diagrams. Our graphical representation, and various transformations of the original graph into a tree structure, are then used to guide fast routines for the computation of a decision problem's expected utilities. We show that our routines generalize those usually utilized in standard influence diagrams' evaluations under much more restrictive conditions. We then proceed with the construction of a directed expected utility network to support decision makers in the domain of household food security. 
\end{abstract}


\end{frontmatter}

\section{Introduction}
The Bayesian paradigm provides a coherent platform to frame the beliefs and the preferences of decision makers (DMs). Once a DM has specified these in the form of a probability distribution and a utility function, then under the subjective expected utility paradigm she would act rationally by choosing a decision that maximizes her expected utility, i.e. the expectation of the utility function with respect to the probability distribution elicited from her. Although other paradigms expressing different canons of rationality exist \citep[e.g.][]{Giang2005, Hong2000, Smets2002}, applied decision making problems have been most commonly addressed within this Bayesian framework \citep{Gomez2004,Heckerman1995}.

One of the reasons behind the widespread use of Bayesian methods is the existence of formally justifiably methods that can be used to decompose utility functions and probability distributions into several others, each of which have a smaller dimension than those of a naive representation of the problem. This decomposition offers both computational advantages and more focused decision-making, since the DM only needs to elicit beliefs on small dimensional subsets of variables. This in turn has led to larger and larger problems being successfully and accurately  modelled within this Bayesian framework.

The decomposition of the probabilistic part of the world is usually achieved via the notion of conditional independence \citep{Dawid1979}. It was long ago recognized that graphical representations of the relationships between random variables directly express a collection of conditional independences. These independences enabled large dimensional joint probabilities to be formally written as products of local distributions of smaller dimension, needing many fewer probability specifications than a direct, full specification. Many formal statistical graphical models were subsequently defined, most notably Bayesian networks (BNs) \citep{Pearl1988,Smith2010}, that exploited these conditional independences to represent the qualitative structure of a multivariate random vector through a directed graph.

There are also many independence concepts related to utility that can be used to factorize a utility function into terms with a smaller number of arguments. Standard independence concepts are based on the notion of (generalized) additive independence and (conditional) utility independence \citep{Keeney1993}, These both entail some additive or multiplicative decomposition of the utility function. Fairly recently it has been recognized that sets of such statements could also be represented by a graph, which in turn could be used to develop fast elicitation routines \citep[see e.g.][]{Abbas2005,Abbas2009,Abbas2010,Abbas2011,GAI,Engel2008,Gonzales2004}.

The class of influence diagrams \citep{Howard2005, Nielsen2009, Smith2008} was one of the first graphical methods to contemporaneously depict probabilistic dependence, the form of the utility function and the structure of the underlying decision space. Fast routines to compute expected utilities and identify optimal decisions that exploit the underlying graph have been defined for a long while \citep[e.g.][]{Jensen1994, Shachter1986}. However, these are almost exclusively designed to work when the utility can be assumed to factorize additively, i.e. assuming that the utility can be written as a linear combination of smaller dimensional functions over disjoint subsets of the decision problem's attributes. An exception is the multiplicative influence diagram \citep{Leonelli2015a}, whose evaluation algorithm works not only for additive factorizations but also for more general multiplicative ones \citep{Keeney1974}.

In this paper we develop a class of graphical models that can depict both probabilistic independence and sets of (conditional) utility independence statements expressible by a utility diagram \citep{Abbas2010}. We call these \textit{directed expected utility networks} (DEUNs). We here develop two fast algorithms for the computation of expected utilities using these diagrams. The first one applies to any DEUN and consists of a sequential application of a conditional expectation operator, analogous to the chance node removal of \citet{Shachter1986}. The second algorithm is valid only for a subset of DEUNs, ones that we call here \textit{decomposable}. After a transformation into a new junction tree representation of the problem, this routine computes the overall expected utility via variable elimination just as in \citet{Jensen1994}, but now applied to our much more general family of utilities. We are able to demonstrate that the elimination step in DEUNs almost exactly coincides with that of standard ID's evaluation algorithms. Therefore both additional theoretical results, as for example approximated propagation, and code already available for IDs, designed originally for use with additive utilities, can be fairly straightforwardly generalized to be used in conjunction with a much more general utility structure.

The motivation for this work stems from a decision support system we are currently building to help local authorities evaluate the impacts of different policies in the light of endemic food poverty \citep{Smith2015,Smith2015a}. In the initial study of \citet{Barons2017} - to keep the analysis as simple as possible - the underlying preferential structure was assumed to factorize additively as commonly made in ID modelling and many applied decision analyses. Discussions during the elicitation process however showed that this assumptions was far from ideal in this application. Currently available technology would not enable us to formally perform a decision analysis under the required much milder preferential conditions. We have thus take on this challenge and developed new algorithms for the computation of expected utilities that enable decision makers to perform much more general decision analyses.

 The only other attempt in the literature we are aware of to represent utility and probabilistic dependence in a unique graph is the expected utility network of \citet{Mura99}. This is an undirected graphical model with two types of edges to represent probabilistic and preferential dependence. However, this method is built on a non-standard notion of a conditional utility function. Furthermore, fast routines for the computation of the associated expected utility have yet been developed using this framework. In contrast, DEUNs are based on commonly used concepts of utility independences characterised by various preference relationships and so directly apply to standard formulations of decision problems.

The paper is structured as follows. In Section \ref{section2} we review the Bayesian paradigm for decision making. In Sections \ref{section3} and \ref{section4} we review independence concepts and their graphical representations for probabilities and utilities, respectively. In Section \ref{section5} we define our DEUN graphical model and in Section \ref{section6} we develop algorithms for the computation of the DEUN's expected utilities. Section \ref{section8} presents an application of DEUNs to household food security. We conclude in Section \ref{section9} with a discussion.

\section{Bayesian decision making}
\label{section2}
Let $d$ be a decision within some set $\mathbb{D}$ of available decisions, $n\in\mathbb{N}$ and $[n]=\{1,\dots,n\}$. Let $\bm{Y}=(Y_i)_{i\in[n]}$ be an absolutely continuous random vector including the attributes of the problem, i.e. the arguments over which a utility function $u$ is defined. For a subset $A\subseteq[n]$, we let $\bm{Y}_A=(Y_i)_{i\in A}$, $\mathbb{Y}_A=\times_{i\in A}\mathbb{Y}_i$, where $\mathbb{Y}_i$ is the sample space of $Y_i$, and denote with $y_i$ and $\bm{y}_A$ instantiations of $Y_i$ and $\bm{Y}_A$, respectively,  $i\in[n]$. Lastly, let $\bm{y}_{[n]}=\bm{y}$ and $\mathbb{Y}_{[n]}=\mathbb{Y}$.

 In this paper, we assume the utility function $u$ to be continuous and normalized between zero and one so that $u:\mathbb{Y}\times \mathbb{D}\rightarrow [0,1]$. In addition, we assume that for each attribute $Y_i$ there are two reference values $y_i^0,y_i^*\in\mathbb{Y}_i$ such that $u(y_i^*,\bm{y}_{-i},d)>u(y_i^0,\bm{y}_{-i},d)$ for every $d\in\mathbb{D}$, where, for a set $A\subset[n]$, $\bm{y}_{-A}=(y_j)_{j\in [n]\setminus A}$.

The expected utility $\overline{u}(d)$ of a decision $d\in\mathbb{D}$ - the expectation of $u(\bm{y},d)$ with respect to the probability density $p(\bm{y} | d)$ - is then
\begin{equation}
\label{eq:eu}
\overline{u}(d)=\mathbb{E}(u(\bm{y},d))=\int_{\mathbb{Y}}u(\bm{y},d)p(\bm{y} | d)\textnormal{d} \bm{y}.
\end{equation}
 A rational decision maker would then choose to enact an \textit{optimal} decision $d^*$, where $d^*=\argmax_{d\in\mathbb{D}}\{\overline{u}(d)\}$.
 
  This framework, though conceptually straightforward, can become very challenging to apply in practice. As soon as the number of attributes grows moderately a faithful elicitation of the probability and utility functions becomes prohibitive. In addition to the knowledge issues in eliciting multivariate functions, the computation of the expected utility in equation (\ref{eq:eu}) requires an integration over an arbitrary large space $\mathbb{Y}$ which, again, may become infeasible in high dimensional settings. For these two reasons various additional models and independence conditions have been imposed. We review these types of conditions in the next two sections.   

For ease of notation in the following we leave implicit the dependence of all arguments of functions of interest on the decision $d\in\mathbb{D}$. On one hand we can assume that both the probabilistic and the utility independence structure are invariant to the choice of $d\in\mathbb{D}$. We note that this is an assumption commonly made in standard influence diagram modelling. Now $p(\bm{y}|d)$ and $u(\bm{y},d)$ may well be functions of $d\in\mathbb{D}$ -  we simply assume that the underlying conditional independence structure and preferential independences are shared by all $d\in\mathbb{D}$. But for any finite discrete space $\mathbb{D}$, we could alternatively apply our methods under the more general assumption that, for each $d\in\mathbb{D}$, the DM's problem could be depicted by a possibly different network. We could then apply the theory we develop below to each of these networks in turn and finally optimise over these separate evaluations - albeit more slowly.

\section{Probability factorizations}
\label{section3}
The concept used in probabilistic modelling to simplify density functions is \textit{conditional independence} \citep{Dawid1979}. For three random variables $Y_i$, $Y_j$ and $Y_k$  with strictly positive joint density we say that $Y_i$ is conditional independent of $Y_j$ given $Y_k$, and write $Y_i\independent Y_j|Y_k$, if the conditional density  of $Y_i$ can be written as a function of $Y_i$ and $Y_k$ only, i.e.
$
p(y_i\,|\, y_j,y_k)=p(y_i\,|\,y_k).
$ 
This means that the only information to infer $Y_i$ from $Y_j$ and $Y_k$ is from $Y_k$. 

Sets of conditional independence statements can then be depicted by a graph whose vertices are associated to the random variables of interest. We next briefly introduce some terminology from graph theory and then define one of the most common statistical graphical models, namely the \textit{Bayesian network}.

\subsection{Graph theory}
A directed \textbf{graph} $\Gr$ is a pair $\Gr=(V(\Gr), E(\Gr))$, where $V(\Gr)$ is a finite set of \emph{vertices} and $E(\Gr)$ is a set of ordered  pairs of vertices, called \emph{edges}. A directed \textit{path} of length $m$ from ${i_1}$ to ${i_m}$ in a graph $\Gr$ is a sequence of $m$ vertices such that, for any two consecutive vertices $i_j$ and $i_{j+1}$ in the sequence, $(i_j,i_{j+1})\in E(\Gr)$. If there is a directed path from $i$ to $j$ in $\Gr$ we write $i\rightarrow j$. We use the symbol $i \not\rightarrow j$ if there is no such directed path in $\Gr$. Conversely, an undirected path is a sequence of vertices such that either $(i_j,i_{j+1})\in E(\Gr)$ or $(i_{j+1},i_j)\in E(\Gr)$. A \textit{cycle} is a directed path  with the additional condition that ${i_1}={i_m}$. For $i,j\in V(\Gr)$, we say that $i$ and $j$ are connected if there is an undirected path between $i$ and $j$. A graph $ \Gr$ is \emph{connected} if every pair of vertices $i,j\in V(\Gr)$ are connected. A \emph{directed acyclic graph (DAG)} is a directed graph with no cycles.  For these graph the labelling of the vertices can be constructed, not uniquely, so that $i<j$ if $(i,j)\in E(\Gr)$. 

Now let $\mathcal{G}$ be a DAG. If $(i,j)\in E(\Gr)$ we say that $i$ is a \textit{parent} of $j$ and that $j$ is a \textit{child} of $i$. The set of  parents of $i$ is denoted by $\Pi_i$.   A vertex of a DAG with no children is called \textit{leaf}, whilst a \textit{root} is a vertex with no parents. A DAG is said to be \textit{decomposable} if all pairs of parents of the same child are joined by an edge. A subset $C$ of $V(\Gr)$ is a \textit{clique} of $\Gr$ if any pair $i,j\in C$ is connected by an edge and there is no other $C'\subseteq V(\Gr)$ with the same property such that $C\subset C'$. Let $\Gr$ have $m$ cliques $\{C_1,\dots,C_m\}=\mathcal{C}$ and suppose the elements of $\mathcal{C}$ are ordered according to their indexing. A \textit{separator} $S_i$ of $\Gr$, $i\in[m]\setminus\{1\}$, is defined as $S_i=C_i\cap\cup_{j=1}^{i-1}C_j$. The cliques of $\Gr$ are said to respect the running intersection property if $S_i\subseteq C_j$ for at least one $j<i$, $i\in[m]\setminus\{1\}$.  

\begin{figure}
\begin{center}
    \begin{subfigure}[b]{0.48\textwidth}
        \centering
       \begin{tikzpicture}[node distance=4cm]
\node[draw,circle] (1) at (0,0) {$1$};
\node[draw,circle](5) at (0,2) {$5$};
\node[draw,circle] (3) at (2,0) {$3$};
\node[draw,circle] (2) at (2,2) {$2$};
\node[draw,circle] (4) at (4,2) {$4$};
\draw[->] (1) -- (5);
\draw[->] (1) -- (2);
\draw[->] (1) -- (3);
\draw[->] (2) -- (3);
\draw[->] (2) -- (4);
\end{tikzpicture}
        \caption{\footnotesize{A decomposable, connected DAG.}}
        \label{fig:DAG}
    \end{subfigure}
    \begin{subfigure}[b]{0.48\textwidth}
    \centering
            \begin{tikzpicture}[node distance=4cm]
\node[draw,circle] (1) at (0,1) {$1$};
\node[draw,circle] (3) at (2,0) {$3$};
\node[draw,circle] (2) at (2,2) {$2$};
\draw[->] (1) -- (3);
\draw[->] (1) -- (2);
\end{tikzpicture}
        \caption{\footnotesize{A directed tree.}}   
        \label{fig:dirtree}
    \end{subfigure}
    \end{center}
    \caption{Example of two DAGs.\label{fig:DAGs}}
    \end{figure}
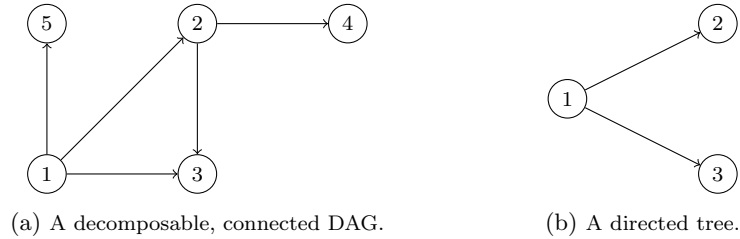

\begin{example}
The directed graph in Figure \ref{fig:DAG} can be clearly seen to be a DAG with vertex set equal to $[5]$. This is decomposable since the two parents of vertex 3, i.e. 1 and 2, are connected by an edge. This DAG is also connected since every two vertices are connected by an undirected path. The cliques of the DAG in Figure \ref{fig:DAG} are $C_1=\{1,2,3\}$, $C_2=\{2,4\}$ and $C_3=\{1,5\}$ and its separators $S_2=\{2\}$  and $S_3=\{1\}$. So with this indexing the cliques of this DAG respects the running intersection property.
\end{example}

A graph of interest in this paper is the \emph{directed tree} $\mathcal{T}$. This is a DAG with the following two properties: it has a unique vertex with no parents called \textit{root}; and all other vertices have exactly one parent. The DAG in Figure \ref{fig:dirtree} can be clearly seen to be a directed tree with root $1$ and leaves $2$ and $3$.

\subsection{Bayesian networks}
We are now ready to define the statistical graphical model that underpins the probabilistic part of the DEUN model we define below.
\begin{definition}
A BN over a random vector $\bm{Y}=(Y_i)_{i\in[n]}$ consists of 
\begin{itemize}
\item $n-1$ \textit{conditional independence} statements of the form $Y_i\independent \bm{Y}_{[i-1]\setminus \Pi_i}\;|\, \bm{Y}_{\Pi_i}$, where $\Pi_i\subseteq [i-1]$;
\item a DAG $\mathcal{\Gr}$ with vertex set $V(\Gr)=[n]$ and edge set $E(\Gr)=\{(i,j):j\in[n],i\in\Pi_j\}$;
\item conditional distributions  $p(y_i\,|\,\bm{y}_{\Pi_i})$ for $i\in[n]$. 
\end{itemize}
\end{definition}
It can be shown \citep[e.g.][]{Lauritzen1996} that the density of a BN can then be written as
\[
p(\bm{y})=\prod_{i\in[n]}p(y_i\,|\,\bm{y}_{\Pi_i}).
\]

\begin{example}
Consider the DAG in Figure \ref{fig:DAG}. A BN with this associated graph implies the conditional independences $Y_4\independent (Y_1,Y_3)\,|\, Y_2$  and  $Y_5\independent (Y_2,Y_3,Y_4)\,|\,Y_1$. The probability distribution then factorizes as
\[
p(\bm{y})=p(y_5\,|\,y_1)p(y_4\,|\,y_2)p(y_3\,|\,y_1,y_2)p(y_2\,|\,y_1)p(y_1).
\]
\end{example}

\section{Utility factorizations}
\label{section4}
\subsection{Independence and factorizations}
Whilst conditional independence is universally acknowledged as the gold standard to simplify probabilistic joint densities, for utility functions a variety of independence concepts have been used. One very common assumption is that a utility has additively independent attributes implying the additive utility factorization
\begin{equation}
\label{eq:add}
u(\bm{y})=\sum_{i\in[n]}k_iu(y_i),
\end{equation}
where $k_i=u(y_i^*,\bm{y}_{-i}^0)$ is a \textit{criterion weight} and $u(y_i)=u(y_i,\bm{y}_{-i}^*)=u(y_i,\bm{y}_{-i}^0)$, $i\in[n]$. A generalization of this independence concept applies to subsets of $[n]$ that are possibly non-disjoint \citep{GAI,Fishburn1967}.

A second approach for defining multivariate utility factorizations is to first identify \textit{utility independences}. For this purpose we introduce the \textit{conditional utility function} of $\bm{y}_A$ given $\bm{y}_{-A}$, $A\subset [n]$,
\[
u(\bm{y}_A\,|\, \bm{y}_{-A})=\frac{u(\bm{y})-u(\bm{y}_A^0,\bm{y}_{- A})}{u(\bm{y}_A^*,\bm{y}_{- A})-u(\bm{y}_A^0,\bm{y}_{- A})},
\]
where $\bm{y}_A^0=(y_i^0)_{i\in A}$ and $\bm{y}_A^{*}=(y_i^*)_{i\in A}$.
\begin{definition}
We say that $\bm{Y}_A$ is utility independent of $\bm{Y}_B$ given $\bm{Y}_C$, $\bm{Y}_A\ \mbox{UI}\ \bm{Y}_B| \bm{Y}_C$, for $A\cup B\cup C=[n]$, if and only if we can write
\[
u(\bm{y}_A\,|\, \bm{y}_B,\bm{y}_C)=u(\bm{y}_A\,|\,\bm{y}_C).
\]
\end{definition}
Utility independences then imply joint utility functions that have a simpler form.  Let $A\subseteq [n]$ be a totally ordered set and let, for each $i\in A$, $iP$ and $iF$ be the set of indices that precede and follow $i$ in $A$, respectively. Let $\mathbb{Y}_A^{0*}$ be the set comprising all possible instantiations of $\bm{Y}_A$, where each element is either $y_i^0$ or $y_i^*$, $i\in A$, and let $\bm{y}_A^{0*}$ be an element of $\mathbb{Y}_A^{0*}$. \citet{Abbas2010} showed that, by sequentially applying conditional utility independence statements according to the order of the elements in $A$, any utility function can then be written as
\begin{equation}
u(\bm{y})=\sum_{\bm{y}_A^{0*}\in\mathbb{Y}_A^{0*}}u(\bm{y}_A^{0*},\bm{y}_{- A})\prod_{i\in A}g(y_i\,|\,\bm{y}_{iP}^{0*},\bm{y}_{iF}),
\label{eq:expansion}
\end{equation}
where 
\[
g(y_i\,|\,\bm{y}_{iP}^{0*},\bm{y}_{iF})=\left\{
\begin{array}{lccl}
u(y_i\,|\,\bm{y}_{iP}^{0*},\bm{y}_{iF}),&&& \mbox{if } y_i=y_i^* \mbox{ in } u(\bm{y}_A^{0*},\bm{y}_{- A}),\\
\hat{u}(y_i\,|\,\bm{y}_{iP}^{0*},\bm{y}_{iF}), &&& \mbox{otherwise},
\end{array}
\right.
\]
and $\hat{u}(y_i\,|\,\bm{y}_{iP}^{0*},\bm{y}_{iF})=1-u(y_i\,|\,\bm{y}_{iP}^{0*},\bm{y}_{iF})$ is the \textit{disutility function}. So for example if each $Y_i$ is utility independent of $\bm{Y}_{-i}$ then equation (\ref{eq:expansion}) can be re-expressed as
\begin{equation}
u(\bm{y})=\sum_{\bm{y}^{0*}\in\mathbb{Y}^{0*}}u(\bm{y}^{0*})\prod_{i\in [n]}g(y_i|\bm{y}_{-i}^0).
\label{eq:multilinear}
\end{equation}
 This special case can be identified as the well-known multilinear utility factorization \citep{Keeney1993}.

\subsection{Utility diagrams}
Graphical models depicting various types of preferential independences have now begun to appear.
In this paper we consider a specific class of models called \textit{utility diagrams} \citep{Abbas2010}. 
\begin{definition}
\label{def:utdiag}
A \textit{utility diagram} is a directed graph with vertex set $[n]$ and its edge set is such that the absence of an edge $(i,j)$, $i,j\in[n]$, implies $Y_j\ \mbox{UI}\ Y_i\ |\ \bm{Y}_{-ij}$.
\end{definition}
Note that \citet{Abbas2010} defined utility diagrams as bidirectional graphs. However, given that our definition of a directed graph allows vertices to be connected by more than one edge, the model in Definition \ref{def:utdiag} is equivalent to the one of \citet{Abbas2010}, where a bidirected edge between two vertices is replaced by two edges, one pointing in each direction.

A utility diagram with empty edge set corresponds to a multilinear factorization of the utility function as in equation (\ref{eq:multilinear}). Here we introduce a subclass of utility diagrams that has some important properties.
\begin{definition}
A utility diagram is said to be \emph{directional} if its graph is a DAG.
\end{definition}

\begin{example}
The utility diagram in Figure \ref{fig:utdiag} is directional and implies the following conditional utility independences
\[
\begin{array}{ccccccc}
Y_1\ \mbox{UI}\ Y_2|Y_3,Y_4,Y_5,&&&Y_1\ \mbox{UI}\ Y_3|Y_2,Y_4,Y_5, &&&Y_1\ \mbox{UI}\ Y_4|Y_2,Y_3,Y_5,\\ Y_1\ \mbox{UI}\ Y_5|Y_2,Y_3,Y_4,&&&Y_2\ \mbox{UI}\ Y_3|Y_1,Y_4,Y_5, &&&Y_2\ \mbox{UI}\ Y_4|Y_1,Y_3,Y_5,\\
Y_2\ \mbox{UI}\ Y_5|Y_1,Y_3,Y_4,&&&Y_3\ \mbox{UI}\ Y_2|Y_1,Y_4,Y_5, &&&Y_3\ \mbox{UI}\ Y_4|Y_1,Y_2,Y_5,\\ Y_3\ \mbox{UI}\ Y_5|Y_1,Y_2,Y_4,&&&Y_4\ \mbox{UI}\ Y_3|Y_1,Y_2,Y_5, &&&Y_4\ \mbox{UI}\ Y_5|Y_1,Y_2,Y_3,\\
Y_5\ \mbox{UI}\ Y_2|Y_1,Y_3,Y_4,&&&Y_5\ \mbox{UI}\ Y_3|Y_1,Y_2,Y_4, &&&Y_5\ \mbox{UI}\ Y_4|Y_1,Y_2,Y_3.
\end{array}
\]

\begin{figure}
\begin{center}
\begin{tikzpicture}[node distance=4cm]
\node[draw,circle] (1) at (0,0) {$1$};
\node[draw,circle](5) at (0,2) {$5$};
\node[draw,circle] (3) at (2,0) {$3$};
\node[draw,circle] (2) at (2,2) {$2$};
\node[draw,circle] (4) at (4,2) {$4$};
\draw[->,sloped, dashed] (1) to  (5);
\draw[->,sloped, dashed] (1) to  (2);
\draw[->,sloped, dashed] (1) to  (3);
\draw[->,sloped, dashed] (1) to  (4);
\draw[->,sloped, dashed] (2) to  (4);
\end{tikzpicture}
\end{center}
\caption{Example of a directional utility diagram. \label{fig:utdiag}}
\end{figure}
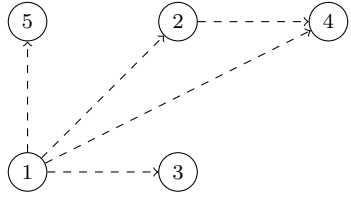

\end{example}

 Directional utility diagrams have the unique property that their  utility function can be written in terms of criterion weights and univariate utility functions only.  Although not explicitly depicted by a utility graph, such a property underlies the algorithms developed in \citet{Leonelli2015} that apply to some specific generalized additively independent models only. 
 
\begin{lemma}
\label{lemma:ut}
For a directional utility diagram there exists an expansion order over $[n]$ such that equation  (\ref{eq:expansion}) is a linear combination of terms involving only criterion weights and conditional utility functions having as argument a single attribute. 
\end{lemma}

This result follows by observing that the terms $u(\bm{y}_I^{0*},\bm{y}_{- I})$ in equation (\ref{eq:expansion}) coincide with $u(\bm{y}^{0*})$ since the expansion can be performed over all the attributes. These terms are functions of criterion weights. Furthermore the conditional independence structure underlying a directed utility diagram is such that there is an expansion order where $\bm{Y}_{i}$ UI $Y_{iF}\;|\; \bm{Y}_{iP}$. Thus $g(y_i\;|\; \bm{y}_{iP}^{0*},\bm{y}_{iF})$ in equation (\ref{eq:expansion}) is equal to $g(y_i\;|\; \bm{y}_{iP}^{0*})$ for every $i\in[n]$.

\begin{example}
The utility factorization associated to the diagram in Figure \ref{fig:utdiag} is equal to the sum of the entries of Table \ref{table}, where $r_I$, $I\subseteq[n]$, is equal to $u(\bm{y}_I^*,\bm{y}_{- I}^0)$. This in general consists of $2^n$ terms each made of $n+1$ indeterminates, where $n$ is number of vertices in the diagram.
\end{example}
\begin{table}
\begin{center}
\scalebox{0.9}{
\begin{tabular}{|c|c|}
\hline
$r_{\emptyset}\hat{u}(y_1)\hat{u}(y_2|y_1^0)\hat{u}(y_3|y_1^0)\hat{u}(y_4|y_1^0,y_2^0)\hat{u}(y_5|y_1^0)$&$r_1u(y_1)\hat{u}(y_2|y_1^*)\hat{u}(y_3|y_1^*)\hat{u}(y_4|y_1^*,y_2^0)\hat{u}(y_5|y_1^*)$\\
$r_{2}\hat{u}(y_1)u(y_2|y_1^0)\hat{u}(y_3|y_1^0)\hat{u}(y_4|y_1^0,y_2^*)\hat{u}(y_5|y_1^0)$&$r_3\hat{u}(y_1)\hat{u}(y_2|y_1^0)u(y_3|y_1^0)\hat{u}(y_4|y_1^0,y_2^0)\hat{u}(y_5|y_1^0)$\\
$r_{4}\hat{u}(y_1)\hat{u}(y_2|y_1^0)\hat{u}(y_3|y_1^0)u(y_4|y_1^0,y_2^0)\hat{u}(y_5|y_1^0)$&$r_{5}\hat{u}(y_1)\hat{u}(y_2|y_1^0)\hat{u}(y_3|y_1^0)\hat{u}(y_4|y_1^0,y_2^0)u(y_5|y_1^0)$\\
$r_{12}u(y_1)u(y_2|y_1^*)\hat{u}(y_3|y_1^*)\hat{u}(y_4|y_1^*,y_2^*)\hat{u}(y_5|y_1^*)$&$r_{13}u(y_1)\hat{u}(y_2|y_1^*)u(y_3|y_1^*)\hat{u}(y_4|y_1^*,y_2^0)\hat{u}(y_5|y_1^*)$\\
$r_{14}u(y_1)\hat{u}(y_2|y_1^*)\hat{u}(y_3|y_1^*)u(y_4|y_1^*,y_2^0)\hat{u}(y_5|y_1^*)$&$r_{15}u(y_1)\hat{u}(y_2|y_1^*)\hat{u}(y_3|y_1^*)\hat{u}(y_4|y_1^*,y_2^0)u(y_5|y_1^*)$\\
$r_{23}\hat{u}(y_1)u(y_2|y_1^0)u(y_3|y_1^0)\hat{u}(y_4|y_1^0,y_2^*)\hat{u}(y_5|y_1^0)$&$r_{24}\hat{u}(y_1)u(y_2|y_1^0)\hat{u}(y_3|y_1^0)u(y_4|y_1^0,y_2^*)\hat{u}(y_5|y_1^0)$\\
$r_{25}\hat{u}(y_1)u(y_2|y_1^0)\hat{u}(y_3|y_1^0)\hat{u}(y_4|y_1^0,y_2^*)u(y_5|y_1^0)$&$r_{34}\hat{u}(y_1)\hat{u}(y_2|y_1^0)u(y_3|y_1^0)u(y_4|y_1^0,y_2^0)\hat{u}(y_5|y_1^0)$\\
$r_{35}\hat{u}(y_1)\hat{u}(y_2|y_1^0)u(y_3|y_1^0)\hat{u}(y_4|y_1^0,y_2^0)u(y_5|y_1^0)$&$r_{45}\hat{u}(y_1)\hat{u}(y_2|y_1^0)\hat{u}(y_3|y_1^0)u(y_4|y_1^0,y_2^0)u(y_5|y_1^0)$\\
$r_{123}u(y_1)u(y_2|y_1^*)u(y_3|y_1^*)\hat{u}(y_4|y_1^*,y_2^*)\hat{u}(y_5|y_1^*)$&$r_{124}u(y_1)u(y_2|y_1^*)\hat{u}(y_3|y_1^*)u(y_4|y_1^*,y_2^*)\hat{u}(y_5|y_1^*)$\\
$r_{125}u(y_1)u(y_2|y_1^*)\hat{u}(y_3|y_1^*)\hat{u}(y_4|y_1^*,y_2^*)u(y_5|y_1^*)$&$r_{134}u(y_1)\hat{u}(y_2|y_1^*)u(y_3|y_1^*)u(y_4|y_1^*,y_2^0)\hat{u}(y_5|y_1^*)$\\
$r_{135}u(y_1)\hat{u}(y_2|y_1^*)u(y_3|y_1^*)\hat{u}(y_4|y_1^*,y_2^0)u(y_5|y_1^*)$&$r_{145}u(y_1)\hat{u}(y_2|y_1^*)\hat{u}(y_3|y_1^*)u(y_4|y_1^*,y_2^0)u(y_5|y_1^*)$\\
$r_{234}\hat{u}(y_1)u(y_2|y_1^0)u(y_3|y_1^0)u(y_4|y_1^0,y_2^*)\hat{u}(y_5|y_1^0)$&$r_{235}\hat{u}(y_1)u(y_2|y_1^0)u(y_3|y_1^0)\hat{u}(y_4|y_1^0,y_2^*)u(y_5|y_1^0)$\\
$r_{245}\hat{u}(y_1)u(y_2|y_1^0)\hat{u}(y_3|y_1^0)u(y_4|y_1^0,y_2^*)u(y_5|y_1^0)$&$r_{345}\hat{u}(y_1)\hat{u}(y_2|y_1^0)u(y_3|y_1^0)u(y_4|y_1^0,y_2^0)u(y_5|y_1^0)$\\
$r_{1234}u(y_1)u(y_2|y_1^*)u(y_3|y_1^*)u(y_4|y_1^*,y_2^*)\hat{u}(y_5|y_1^*)$&$r_{1235}u(y_1)u(y_2|y_1^*)u(y_3|y_1^*)\hat{u}(y_4|y_1^*,y_2^*)u(y_5|y_1^*)$\\
$r_{1245}u(y_1)u(y_2|y_1^*)\hat{u}(y_3|y_1^*)u(y_4|y_1^*,y_2^*)u(y_5|y_1^*)$&$r_{1345}u(y_1)\hat{u}(y_2|y_1^*)u(y_3|y_1^*)u(y_4|y_1^*,y_2^0)u(y_5|y_1^*)$\\
$r_{2345}\hat{u}(y_1)u(y_2|y_1^0)u(y_3|y_1^0)u(y_4|y_1^0,y_2^*)u(y_5|y_1^0)$&$r_{12345}u(y_1)u(y_2|y_1^*)u(y_3|y_1^*)u(y_4|y_1^*,y_2^*)u(y_5|y_1^*)$\\
\hline
\end{tabular}}
\end{center}
\caption{Terms in the utility expansion associated to the utility diagram in Figure \ref{fig:utdiag}. \label{table}}
\end{table}

Focusing on the subclass of directed utility diagrams has the great computational advantage of allowing for the computation of the expected utility of a DEUN through a backward inductive routine. At each step this  computes a finite number of integrals over the sample space of one random variable only. More general utility dependence structures could also be studied by extending our methods: see Section \ref{section9} for a discussion. However, for simplicity in this paper we restrict ourselves to this special case.

\section{Directed expected utility networks}
\label{section5}

We are now ready to define our graphical model which embeds both probabilistic and utility independence statements. 
\begin{definition}
A \textbf{directed expected utility network} $\mathcal{G} $ consists of a set of vertices $V(\mathcal{G})=[n]$, a probabilistic edge set $E_p(\mathcal{G})$, denoted by solid arrows, and a utility edge set $E_u(\mathcal{G})$, denoted by dashed arrows, such that:
\begin{itemize}
\item $(V(\mathcal{G}),E_p(\mathcal{G}))$ is a BN model such that if $(i,j)\in E_p(\mathcal{G})$ then $i<j$;
\item $(V(\mathcal{G}), E_u(\mathcal{G}))$ is a directional utility diagram such that if $(i,j)\in E_u(\mathcal{G})$ then $i<j$.
\end{itemize}
\end{definition} 

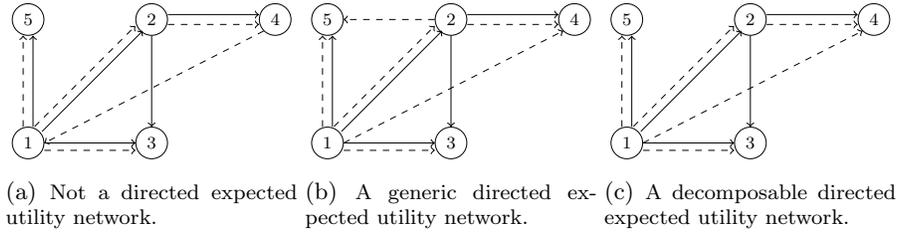
\begin{figure}
\begin{center}
    \begin{subfigure}[b]{0.32\textwidth}
        \centering
        \resizebox{\linewidth}{!}{
        \begin{tikzpicture}[node distance=4cm]
\node[draw,circle] (1) at (0,0) {$1$};
\node[draw,circle](5) at (0,2) {$5$};
\node[draw,circle] (3) at (2,0) {$3$};
\node[draw,circle] (2) at (2,2) {$2$};
\node[draw,circle] (4) at (4,2) {$4$};
\draw[->,sloped, anchor=center, above,near start] ([xshift=0.08cm]1.north) to  node {} ([xshift=0.08cm]5.south);
\draw[->,dashed,sloped, anchor=center, below,near start] ([xshift=-0.08cm]1.north) to  node {} ([xshift=-0.08cm]5.south);
\draw[->,sloped, anchor=center, above,near start] ([xshift=0.04cm]1.north east) to  node {} ([xshift=0.04cm]2.south west);
\draw[->,dashed,sloped, anchor=center, below,near start] ([xshift=-0.08cm,yshift=0.08cm]1.north east) to  node {} ([xshift=-0.08cm,yshift=0.08cm]2.south west);
\draw[->,dashed,sloped, anchor=center, below,near start] (4.south west) to  node {} (1.east);
\draw[->] (1) -- (3);
\draw[->,dashed,sloped, anchor=center, below,near start] ([yshift=-0.12cm]1.east) to  node {} ([yshift=-0.12cm]3.west);
\draw[->] (2) -- (3);
\draw[->,sloped, anchor=center, above,near start] ([yshift=0.08cm]2.east) to  node {} ([yshift=0.08cm]4.west);
\draw[->,dashed,sloped, anchor=center, below,near start] ([yshift=-0.08cm]2.east) to  node {} ([yshift=-0.08cm]4.west);
\end{tikzpicture}    
        }
        \caption{\footnotesize{Not a directed expected utility network.}}
        \label{fig:subfig8}
    \end{subfigure}
    \begin{subfigure}[b]{0.32\textwidth}
    \centering
        \resizebox{\linewidth}{!}{
            \begin{tikzpicture}[node distance=4cm]
\node[draw,circle] (1) at (0,0) {$1$};
\node[draw,circle](5) at (0,2) {$5$};
\node[draw,circle] (3) at (2,0) {$3$};
\node[draw,circle] (2) at (2,2) {$2$};
\node[draw,circle] (4) at (4,2) {$4$};
\draw[->,sloped, anchor=center, above,near start] ([xshift=0.08cm]1.north) to  node {} ([xshift=0.08cm]5.south);
\draw[->,dashed,sloped, anchor=center, below,near start] ([xshift=-0.08cm]1.north) to  node {} ([xshift=-0.08cm]5.south);
\draw[->,sloped, anchor=center, above,near start] ([xshift=0.04cm]1.north east) to  node {} ([xshift=0.04cm]2.south west);
\draw[->,dashed,sloped, anchor=center, below,near start] ([xshift=-0.08cm,yshift=0.08cm]1.north east) to  node {} ([xshift=-0.08cm,yshift=0.08cm]2.south west);
\draw[->,dashed,sloped, anchor=center, below,near start] (1.east) to  node {} (4.south west);
\draw[->] (1) -- (3);
\draw[->,dashed,sloped, anchor=center, below,near start] ([yshift=-0.12cm]1.east) to  node {} ([yshift=-0.12cm]3.west);
\draw[->] (2) -- (3);
\draw[->,dashed] (2) -- (5);
\draw[->,sloped, anchor=center, above,near start] ([yshift=0.08cm]2.east) to  node {} ([yshift=0.08cm]4.west);
\draw[->,dashed,sloped, anchor=center, below,near start] ([yshift=-0.08cm]2.east) to  node {} ([yshift=-0.08cm]4.west);
\end{tikzpicture}
        }
        \caption{\footnotesize{A generic directed expected utility network.}}   
        \label{fig:subfig9}
    \end{subfigure}
    \begin{subfigure}[b]{0.32\textwidth}
        \centering
        \resizebox{\linewidth}{!}{
            \begin{tikzpicture}[node distance=4cm]
\node[draw,circle] (1) at (0,0) {$1$};
\node[draw,circle](5) at (0,2) {$5$};
\node[draw,circle] (3) at (2,0) {$3$};
\node[draw,circle] (2) at (2,2) {$2$};
\node[draw,circle] (4) at (4,2) {$4$};
\draw[->,sloped, anchor=center, above,near start] ([xshift=0.08cm]1.north) to  node {} ([xshift=0.08cm]5.south);
\draw[->,dashed,sloped, anchor=center, below,near start] ([xshift=-0.08cm]1.north) to  node {} ([xshift=-0.08cm]5.south);
\draw[->,sloped, anchor=center, above,near start] ([xshift=0.04cm]1.north east) to  node {} ([xshift=0.04cm]2.south west);
\draw[->,dashed,sloped, anchor=center, below,near start] ([xshift=-0.08cm,yshift=0.08cm]1.north east) to  node {} ([xshift=-0.08cm,yshift=0.08cm]2.south west);
\draw[->,dashed,sloped, anchor=center, below,near start] (1.east) to  node {} (4.south west);
\draw[->] (1) -- (3);
\draw[->,dashed,sloped, anchor=center, below,near start] ([yshift=-0.12cm]1.east) to  node {} ([yshift=-0.12cm]3.west);
\draw[->] (2) -- (3);
\draw[->,sloped, anchor=center, above,near start] ([yshift=0.08cm]2.east) to  node {} ([yshift=0.08cm]4.west);
\draw[->,dashed,sloped, anchor=center, below,near start] ([yshift=-0.08cm]2.east) to  node {} ([yshift=-0.08cm]4.west);
\end{tikzpicture}
        }
        \caption{\footnotesize{A decomposable directed expected utility network.}}
        \label{fig:subfig10}
    \end{subfigure}
    \end{center}
\caption{Graphical representions of probabilistic and utility independences.} 
\label{fig:DEUN}
\end{figure}

\begin{example}
Consider the diagrams in Figure \ref{fig:DEUN}. Figure \ref{fig:subfig8} includes a graph which is not a DEUN since there is a utility edge from $4$ to $1$. This edge would make the computation of expected utilities via backward induction impossible. Figures \ref{fig:subfig9} and \ref{fig:subfig10} are DEUNs since for these $(V(\mathcal{G}),E_p(\mathcal{G}))$ is a BN and $(V(\mathcal{G}), E_u(\mathcal{G}))$ is a directed utility diagram both including only edges $(i,j)$ such that $i<j$. Note that all three diagrams embed the BN in Figure \ref{fig:DAG}, whilst only the diagram in Figure \ref{fig:subfig10} embeds the utility diagram in Figure \ref{fig:utdiag}.
\end{example}

Note that a DEUN is not allowed to contain any cyclical structure in the edge set of the utility diagram. This is because such cycles would inhibit the computation of expected utility through a backward induction procedure where each node is considered individually and sequentially. Of course it may well be possible to develop more general algorithms by merging the vertices that are connected by such a cycle into a single chain component. However,  the extended flexibility of having two different edge sets would then  need to be offset against the potential loss of both structural information and computational speed. 

We next introduce a subclass of DEUNs that entail fast computation routines. 
\begin{definition}
A DEUN is said to be \textbf{decomposable} if 
\begin{itemize}
\item $(V(\mathcal{G}),E_p(\mathcal{G}))$ is decomposable;
\item $(i,j)\in E_u(\mathcal{G})$ only if $i\rightarrow j$ in  $(V(\mathcal{G}),E_p(\mathcal{G}))$.
\end{itemize}
\end{definition}

\begin{example}
The DEUN in Figure \ref{fig:subfig9} is not decomposable since $(2,5)\in E_u(\mathcal{G})$ but these two vertices are not connected by a directed path in the underlying BN. Conversely the network in Figure \ref{fig:subfig10} is decomposable. Note that  the semantics of our model permit two vertices to be connected by both probabilistic and utility edges, by just one of the two, or potentially none. So for example $(1,2)\in E_p(\mathcal{G})$ and $(1,2)\in E_u(\mathcal{G})$, whilst $(1,4)\not\in E_p(\mathcal{G})$ and $(1,4)\in E_u(\mathcal{G})$. 
\end{example}

Just as in the triangulation step for probabilistic propagation \citep[e.g.][]{Lauritzen1996}, it can be fairly easily showed that any non-decomposable DEUN can be transformed into a decomposable one.
\begin{proposition}
Let $\mathcal{G}$ be a non-decomposable DEUN with vertex set $V(\mathcal{G})$ and edges $E_u(\mathcal{G})$ and $E_p(\mathcal{G})$. Let $\mathcal{G}'$ be a DEUN with vertex set $V(\mathcal{G}')=V(\mathcal{G})$ and edges $E_u(\mathcal{G}')=E_u(\mathcal{G})$ and $E_p(\mathcal{G}')=E_p(\mathcal{G})\cup B_1\cup B_2$, where
\begin{align*}
B_1&=\{(i,j)\in E_u(\mathcal{G}): i\not\rightarrow j \mbox{ in } (V(\mathcal{G}),E_p(\mathcal{G}))\},\\
 B_2&=\{(i,j): (i,k)\land (j,k)\in E_p(\mathcal{G})\cup B_1, \ \forall\ k\in V(\mathcal{G})\}.
\end{align*}
Then $\mathcal{G}'$ is decomposable.
\end{proposition}
This holds by noting that the set $B_1$ simply adds a probabilistic edge connecting two vertices linked by a utility edge which breaks the decomposability condition. The set $B_2$ then simply transform the graph $(V(\mathcal{G}),E_p(\mathcal{G})\cup B_1)$ into a decomposable DAG.

\begin{example}
For the non-decomposable network in Figure \ref{fig:subfig9}, the decomposability condition is achieved by simply adding $(2,5)$ to $E_p(\mathcal{G})$.
\end{example}

\section{Computation of expected utilities}
\label{section6}
We next consider the computation of expected utilities for both non-decomposable and decomposable DEUNs and define algorithms based on backward inductive routines. All these routines have in common an operation working over vectors of (expected) utility functions that we define next. Let $\Pi_i^{u}$ and $\Pi_i^p$ be the parent sets of $i$ with respect to $E_u(\mathcal{G})$ and $E_p(\mathcal{G})$, respectively. We  let $\bm{u}_i(y_i|\bm{y}^{0*}_{\Pi_i^u})=(u(y_i|y_{\Pi_i^u}^{0*}),\hat{u}(y_i|y_{\Pi_i^u}^{0*}))_{y_{\Pi_i^u}^{0*}\in\mathbb{Y}_{\Pi_i^u}^{0*}}$ be the vector comprising the conditional utilities and disutilities given all possible combinations of the parents at the reference values and $\bm{u}_0(\bm{y}^{0*})=(u(\bm{y}^{0*}))_{\bm{y}^{0*}\in\mathbb{Y}^{0*}}$. 
\begin{example}
\label{ex:8}
The vector $\bm{u}_5(y_5|\bm{y}_{\Pi_5^u}^{0*})$ has as its components
\[
\begin{array}{cccccccccccccc}
u(y_5|y_1^*),&&&& u(y_5|y_1^0),&&&& \hat{u}(y_5|y_1^*),&&&& \hat{u}(y_5|y_1^0),
\end{array}
\]
whilst the vector $\bm{u}_4(y_4|\bm{y}_{\Pi_4^u}^{0*})$ has the utility components
\[
\begin{array}{cccccccccccccc}
u(y_4|y_1^*,y_2^*),&&&& u(y_4|y_1^0,y_2^*),&&&&u(y_4|y_1^*,y_2^0),&&&&u(y_4|y_1^0,y_2^0),\\ \hat{u}(y_4|y_1^*,y_2^*),&&&& \hat{u}(y_4|y_1^0,y_2^*),&&&&\hat{u}(y_4|y_1^*,y_2^0),&&&&\hat{u}(y_4|y_1^0,y_2^0).
\end{array}
\] 
\end{example} 
We next introduce an element-wise operation, denoted by $\circ$, which multiplies an element of one vector, $\bm{u}_i(\cdot)$, with any element of another vector, $\bm{u}_j(\cdot)$, if these have compatible instantiations, i.e. if the common conditioning variables are instantiated to the same value. So in our Example \ref{ex:8}, $\bm{u}_5(y_5|\bm{y}_{\Pi_5^u}^{0*})\circ \bm{u}_4(y_4|\bm{y}_{\Pi_4^u}^{0*})$ returns a vector with elements
\[
\begin{array}{ccccc}
u(y_5|y_1^*)u(y_4|y_1^*,y_2^*),&&&&u(y_5|y_1^*)u(y_4|y_1^*,y_2^0),\\
(y_5|y_1^*)\hat{u}(y_4|y_1^*,y_2^*),&&&&u(y_5|y_1^*)\hat{u}(y_4|y_1^*,y_2^0),\\
\hat{u}(y_5|y_1^*)u(y_4|y_1^*,y_2^*),&&&&\hat{u}(y_5|y_1^*)u(y_4|y_1^*,y_2^0),\\
\hat{u}(y_5|y_1^*)\hat{u}(y_4|y_1^*,y_2^*),&&&&\hat{u}(y_5|y_1^*)\hat{u}(y_4|y_1^*,y_2^0),\\
u(y_5|y_1^0)u(y_4|y_1^0,y_2^*),&&&&u(y_5|y_1^0)u(y_4|y_1^0,y_2^0),\\
u(y_5|y_1^0)\hat{u}(y_4|y_1^0,y_2^*),&&&&u(y_5|y_1^0)\hat{u}(y_4|y_1^0,y_2^0),\\
\hat{u}(y_5|y_1^0)u(y_4|y_1^0,y_2^*),&&&&\hat{u}(y_5|y_1^0)u(y_4|y_1^0,y_2^0),\\
\hat{u}(y_5|y_1^0)\hat{u}(y_4|y_1^0,y_2^*),&&&&\hat{u}(y_5|y_1^0)\hat{u}(y_4|y_1^0,y_2^0).
\end{array}
\]
If the vertices $i$ and $j$ are such that $\Pi_i^u\cap\Pi_j^u=\emptyset$ and $\bm{u}_i(\cdot)$ and $\bm{u}_j(\cdot)$ include, respectively, $n_i$ and $n_j$ elements, then $\bm{u}_i(\cdot)\circ\bm{u}_j(\cdot)$ returns a vector of $n_i\times n_j$ entries consisting of all possible multiplications between elements of the vectors. This operation can be encoded by defining the vectors to have elements appropriately ordered so that the standard element-wise multiplication returns only terms having compatible instantiations, just as in \citet{Leonelli2015a}. 
 
\subsection{Computations in generic directed expected utility networks}
The expected utility associated to any DEUN can now be computed via a backward induction which at each step computes a conditional expectation, just as in the chance node removal step of \citet{Shachter1986}. This is formalized in the following theorem.
\begin{theorem}
\label{theo1}
The expected utility score $\overline{u}$  associated to a DEUN $\mathcal{G}$ can be computed according to the following algorithm:
\begin{enumerate}
\item compute:
\begin{equation}
\label{eq:theo1}
\overline{\bm{u}}_n=\int_{\mathbb{Y}_n}\bm{u}_n(y_n|\bm{y}_{\Pi_n^u}^{0*})p(y_n|\bm{y}_{\Pi_n^p})\textnormal{d} y_n,
\end{equation}
\item for $i$ from $n-1$ to $1$, compute:
\begin{equation}
\label{eq:theo11}
\overline{\bm{u}}_i=\int_{\mathbb{Y}_i}\left(\overline{\bm{u}}_{i+1}\circ\bm{u}_i(y_i|\bm{y}_{\Pi_i^u}^{0*})\right)p(y_i|\bm{y}_{\Pi_i^p})\textnormal{d} y_i,
\end{equation}
\item return:
\begin{equation}
\label{eq:theo12}
\overline{u}=|\bm{u}_0(\bm{y}^{0*})\circ \overline{\bm{u}}_1|,
\end{equation}
where, for a vector $\bm{a}$,  $|\bm{a}|$  denotes the sum of its elements.
\end{enumerate}
\end{theorem}
\proof{Proof.}
Define for $i\in[n]$
\begin{equation*}
\tilde{\bm{u}}_i=\circ_{j\in[i]}\bm{u}_j(y_j|\bm{y}_{\Pi^u_j}),
\end{equation*}
and note that 
\begin{equation}
\label{proof1}
\overline{u}=\int_{\mathbb{Y}}u(\bm{y})p(\bm{y})\textnormal{d} \bm{y}=\left|\bm{u}_0(\bm{y}^{0*})\circ\int_{\mathbb{Y}}\tilde{\bm{u}}_np(\bm{y})\textnormal{d}\bm{y}\right|.
\end{equation}
Now consider the second integral in equation (\ref{proof1}). We have that
\begin{align}
\int_{\mathbb{Y}}\tilde{\bm{u}}_np(\bm{y})\textnormal{d}\bm{y}&= \int_{\mathbb{Y}} (\tilde{\bm{u}}_{n-1}\circ\bm{u}(y_n|\bm{y}^{0*}_{\Pi^u_n}))p(y_n|\bm{y}_{\Pi^p_n})p(\bm{y}_{[n-1]})\textnormal{d}\bm{y}\nonumber\\
&= \int_{\mathbb{Y}_{[n-1]}}\tilde{\bm{u}}_{n-1}p(\bm{y}_{[n-1]})\circ\int_{\mathbb{Y}_n}\bm{u}(y_n|\bm{y}_{\Pi^u_n}^{0*})p(y_n|\bm{y}_{\Pi^p_n})\textnormal{d} y_n\textnormal{d} \bm{y}_{[n-1]}\nonumber\\
&=\int_{\mathbb{Y}_{[n-1]}}\tilde{\bm{u}}_{n-1}p(\bm{y}_{[n-1]})\circ \overline{\bm{u}}_n\textnormal{d} \bm{y}_{[n-1]}\nonumber\\
&=\int_{\mathbb{Y}_{[n-1]}}(\tilde{\bm{u}}_{n-1}\circ\overline{\bm{u}}_n)p(\bm{y}_{[n-1]})\textnormal{d} \bm{y}_{[n-1]},\label{eq:proof12}
\end{align}
where $\overline{\bm{u}}_n$ is defined in equation (\ref{eq:theo1}). By marginalizing out $y_{n-1}$, we can then deduce from equation (\ref{eq:proof12}) that
\begin{align}
\int_{\mathbb{Y}}\tilde{\bm{u}}_np(\bm{y})\textnormal{d}\bm{y}&=\int_{\mathbb{Y}_{[n-2]}}\int_{\mathbb{Y}_{n-1}}\hspace{-0.1cm}(\tilde{\bm{u}}_{n-2}\circ\bm{u}(y_{n-1}|\bm{y}^{0*}_{\Pi^u_{n-1}})\circ \overline{\bm{u}}_n)p(y_{n-1}|\bm{y}_{\Pi^p_{n-1}})p(\bm{y}_{[n-2]})\textnormal{d} y_{n-1}\textnormal{d}\bm{y}_{[n-2]}\label{eq:boh}\\
&=\int_{\mathbb{Y}_{[n-2]}}\tilde{\bm{u}}_{n-2}p(\bm{y}_{[n-2]})\circ\int_{\mathbb{Y}_{n-1}}(\bm{u}(y_{n-1}|\bm{y}^{0*}_{\Pi^u_{n-1}})\circ \overline{\bm{u}}_n)p(y_{n-1}|\bm{y}_{\Pi^p_{n-1}})\textnormal{d} y_{n-1}\textnormal{d}\bm{y}_{[n-2]}.
\end{align}
From equation (\ref{eq:theo11}) of Theorem \ref{theo1}, it then follows that 
\begin{align}
\int_{\mathbb{Y}}\tilde{\bm{u}}_np(\bm{y})\textnormal{d}\bm{y}&=\int_{\mathbb{Y}_{[n-2]}}\tilde{\bm{u}}_{n-2}p(\bm{y}_{[n-2]})\circ \overline{\bm{u}}_{n-1}\textnormal{d}\bm{y}_{[n-2]}\nonumber\\
&=\int_{\mathbb{Y}_{[n-2]}}(\tilde{\bm{u}}_{n-2}\circ\overline{\bm{u}}_{n-1})p(\bm{y}_{[n-2]})\textnormal{d}\bm{y}_{[n-2]}.\label{eq:boh2}
\end{align}
By sequentially repeating the steps in equations (\ref{eq:boh})-(\ref{eq:boh2}), we can now deduce that after the marginalization of $Y_2$
\begin{align}
\int_{\mathbb{Y}}\tilde{\bm{u}}_np(\bm{y})\textnormal{d}\bm{y}&=\int_{\mathbb{Y}_{[1]}}(\tilde{\bm{u}}_{1}\circ\overline{\bm{u}}_2)p(\bm{y}_{[1]})\textnormal{d}\bm{y}_{[1]}=\int_{\mathbb{Y}_1}(\bm{u}(y_1)\circ\overline{\bm{u}}_2)p(y_1)\textnormal{d} y_1=\overline{\bm{u}}_1.\label{eq:bah}
\end{align}
Therefore by plugging in equation (\ref{eq:bah}) into (\ref{proof1}), we can conclude that equation (\ref{eq:theo12}) holds. 
\endproof

The above algorithm can be applied directly to any DEUN and computes expected utilities relatively fast and in a distributed fashion by marginalization of individual random variables. However, we also notice that the speed of such a routine can be improved since the computation and transmission of terms that it uses are not strictly necessary. To see this, consider the network in Figure \ref{fig:subfig10}. The algorithm starts from vertex $5$ and computes a marginalization of $\bm{u}(y_5|y_1)$ with respect to the density $p(y_5|y_1)$. The result of this operation, $\overline{\bm{u}}_5$ is then a function of $y_1$ only. In the  algorithm in Theorem \ref{theo1} $\overline{\bm{u}}_5$ is then passed to $4$ and a marginalization with respect to density $p(y_4|y_2)$ is computed over $\overline{\bm{u}}_5\circ\bm{u}_4(y_4|y_2,y_1)$. But $\overline{\bm{u}}_5$ is not a function of $y_4$ and therefore  does not carry any information about this variable which would need to be formally accounted for during its marginalization. Furthermore, since $\bm{u}_4(y_4|y_2,y_1)$ is a function of not only $y_1$ but also $y_2$, the $\circ$ product computes a potentially very large number of terms that are not relevant at this stage of the evaluation. This inefficiency becomes even larger for non connected networks, since the contribution of each of the components can be collated together at the very end of the evaluation. This is because the only joint information these provide lies in the terms $u(\bm{y}^{0*})$.

\subsection{Computations in decomposable directed expected utility networks}
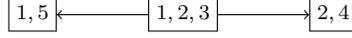
\begin{figure}
\begin{center}
 \begin{tikzpicture}[node distance=4cm]
\node[draw,rectangle] (H) at (0,0) {$1,5$};
\node[draw,rectangle](E) at (2,0) {$1,2,3$};
\node[draw,rectangle] (S) at (4,0) {$2,4$};
\draw[->] (E) -- (H);
\draw[->] (E) -- (S);
\end{tikzpicture}
\end{center}
\caption{Junction tree representation of the DEUN in Figure \ref{fig:subfig10}. \label{fig:jtree}}
\end{figure}

To address these inefficiencies we introduce next a much faster algorithm that works over a transformation of the original graph into a tree structure, just as in standard BNs and IDs junction tree representations \citep[see e.g.][]{Jensen1994}. Let $\mathcal{C}=\{C_i, i\in [m]\}$ be the cliques of the DAG $(V(\mathcal{G}),E_p(\mathcal{G}))$, $S_2,\dots,S_m$ its separators and assume the cliques are ordered to respect the running intersection property. 
\begin{definition}
We call \textit{junction tree} of a decomposable DEUN $\mathcal{G}$ the directed tree $\mathcal{T}$ with vertex set $V(\mathcal{T})=\mathcal{C}$ and edges $(C_i,C_j)$ for one $i\in[j-1]$ such that $S_j\subseteq C_i$, $j\in[m]$. 
\end{definition}
Note that in order to construct such a tree we can straightforwardly apply any of the algorithms already devised for both BNs and IDs \citep[see e.g.][]{Cowell2007}. Furthermore, as for BNs and IDs, a DEUN can have more than one junction tree representation. 
\begin{example}
The junction tree associated to the DEUN in Figure \ref{fig:subfig10} is shown in Figure \ref{fig:jtree}.
\end{example}

In contrast to an algorithm based directly on Theorem \ref{theo1}, we instead propagate using \lq\lq{p}otentials\rq\rq{,} as in many propagation algorithms of BNs and IDs. This enables us to demonstrate that our evaluation algorithm mirrors those commonly used to compute expected utilities in IDs, but now for utility functions that are not necessarily additive. 

Recall that a potential $\Phi_A$, $A\subset [n]$, is a function $\Phi_A:\mathbb{Y}_A\rightarrow \mathbb{R}$. Just as for IDs, we have two types of potentials: utility and probability potentials. For a clique $C\in\mathcal{C}\setminus \{C_1\}$ with an associated separator $S$, its probability potential $\Phi_{C}$ and its utility potential $\Psi_{C}$ are defined as 
\begin{equation*}
\Phi_{C}=\prod_{i\in C\setminus S }p(y_i|\bm{y}_{\Pi_i^p}),\hspace{1cm} \Psi_{C}=\circ_{i\in C\setminus S}\bm{u}(y_i|\bm{y}^{0*}_{\Pi_i^u}), 
\end{equation*}
and $\Phi_{C_1}=\prod_{i\in C_1}=p(y_i|\bm{y}_{\Pi_i^p})$ and $\Psi_{C_1}=\bm{u}(\bm{y}^{0*})\circ_{i\in C_1}\bm{u}(y_i|\bm{y}^{0*}_{\Pi_i^u})$.
Call $\Phi_\mathcal{T}=\prod_{C\in\mathcal{C}}\Phi_{C}$ and $\Psi_{\mathcal{T}}=|\circ_{C\in\mathcal{C}}\Psi_C|$ and note that $p(\bm{y})=\Phi_\mathcal{T}$ and $u(\bm{y})=\Psi_\mathcal{T}$.

Now let $C_i$ be the parent of $C_j$ in $\mathcal{T}$. We say that $C_i$ absorbs $C_j$ if the utility potential of $C_i$, $\Psi_{C_i}$, maps to $\Psi_{C_i}^{C_j}$ where
\begin{equation}
\label{eq:absorption}
\Psi_{C_i}^{C_j}=\Psi_{C_i}\circ \int_{\mathbb{Y}_{C_j\setminus S_j}}\Psi_{C_j}\Phi_{C_j}\textnormal{d} \bm{y}_{C_j\setminus S_j}.
\end{equation}
For a leaf $L$ of $\mathcal{T}$, call $\Phi_{\mathcal{T}\setminus L}$ and $\Psi_{\mathcal{T}\setminus L}$ the probability and utility potentials respectively of the junction tree obtained by absorbing $L$ into its parent and removing $L$ from $\mathcal{T}$. 
\begin{theorem}
\label{theo2}
After absorption of a leaf $L$ with separator $S$ into its parent, we have 
\begin{equation*}
\int_{\mathbb{Y}_{L\setminus S}}\Phi_{\mathcal{T}}\Psi_{\mathcal{T}}\textnormal{d} \bm{y}_{L\setminus S}=\Phi_{\mathcal{T}\setminus L}\Psi_{\mathcal{T}\setminus L}.
\end{equation*}
\end{theorem}
\proof{Proof.} Call
\begin{equation*}
\overline{\Phi}_L=\prod_{C\in\mathcal{C}\setminus L}\Phi_C, \hspace{0.5cm} \mbox{and} \hspace{0.5cm}  \overline{\Psi}_L=\circ_{C\in\mathcal{C}\setminus L}\Psi_C.
\end{equation*}
We have that
\begin{align}
\int_{\mathbb{Y}_{L\setminus S}}\Phi_{\mathcal{T}}\Psi_{\mathcal{T}}\textnormal{d} \bm{y}_{L\setminus S}&=\int_{\mathbb{Y}_{L\setminus S}}\overline{\Phi}_{L}\Phi_L(|\Psi_{L}\circ\overline{\Psi}_L|)\textnormal{d} \bm{y}_{L\setminus S}\nonumber\\
&=\left|\int_{\mathbb{Y}_{L\setminus S}}\overline{\Phi}_{L}\Phi_L(\Psi_{L}\circ\overline{\Psi}_{L\setminus S})\textnormal{d} \bm{y}_{L\setminus S}\right|\nonumber\\
&= \left|\overline{\Phi}_{L}\int_{\mathbb{Y}_{L\setminus S}}(\Phi_L\Psi_{L}\circ\overline{\Psi}_L)\textnormal{d} \bm{y}_{L\setminus S}\right|\nonumber\\
&=\left|\overline{\Phi}_{L}\overline{\Psi}_L\circ\int_{\mathbb{Y}_{L\setminus S}}\Phi_L\Psi_{L}\textnormal{d} \bm{y}_{L\setminus S}\right|.
\label{above}
\end{align}
Writing $\overline{\Psi}_L=\Psi_{\Pi_L}\circ\hat{\Psi}_L$, where $\Psi_{\Pi_L}$ is the utility potential of the parent clique of $L$ and $\hat{\Psi}_L=\circ_{C\in\mathcal{C}\setminus L\setminus \Pi_L}\Psi_C$, it then follows from equation (\ref{above}) that 
\begin{align*}
\int_{\mathbb{Y}_{L\setminus S}}\Phi_{\mathcal{T}}\Psi_{\mathcal{T}}\textnormal{d} \bm{y}_{L\setminus S}&=\left|\overline{\Phi}_L\hat{\Psi}_L\circ\Psi_{\Pi_L}\circ\int_{\mathbb{Y}_{L\setminus S}}\Phi_L\Psi_{L}\textnormal{d} \bm{y}_{L\setminus S}\right|\\
&=\overline{\Phi}_L(|\hat{\Psi}_L\circ\Psi^L_{\Pi_L}|)=\Phi_{\mathcal{T}\setminus L}\Psi_{\mathcal{T}\setminus L}. 
\end{align*}
\endproof

Theorem \ref{theo2} provides the basic step for computing the expected utility of a decomposable DEUN. Suppose the junction tree is connected. Then by sequentially absorbing leaves into parents (for example by following in reverse order the indices of the cliques) we obtain a tree consisting of a vertex/clique only, coinciding with the initial root of the junction tree. Let  $\Psi_{C_1}^{C_2}$ be its probability and utility potentials resulting from the absorption of all the other cliques, assuming $C_2$ was the last clique to be absorbed. It then follows that the overall expected utility $\overline{u}$ is given by
\begin{equation*}
\overline{u}=\int_{\mathbb{Y}_{C_1}}\Phi_{C_1}|\Psi_{C_1}^{C_2}|\textnormal{d} \bm{y}_{C_1}=\left|\int_{\mathbb{Y}_{C_1}}\Phi_{C_1}\Psi_{C_1}^{C_2}\textnormal{d} \bm{y}_{C_1}\right|.
\end{equation*} 

If on the other hand the junction tree is not connected, and this is the case whenever the DEUN is not connected, $\overline{u}$ simply  equals the $\circ$ product of the contributions of the roots of each non-connected components after all other vertices have been absorbed. More formally, let $R_1,\dots,R_k$ be the roots of the non-connected components of the junction tree and let $\Psi_{R_i}^{C_i}$ be their  utility potentials resulting from the absorption of all other cliques, where $C_i$ was the last children of $R_i$ to be absorbed, for $i\in[k]$. We then have that
\begin{equation*}
\overline{u}=\left|\int_{\mathbb{Y}_{R_1}}\Phi_{R_1}^{C_1}\Psi_{R_1}\textnormal{d} \bm{y}_{R_1}\circ\cdots\circ\int_{\mathbb{Y}_{R_k}}\Phi_{R_k}\Psi_{R_k}^{C_k}\textnormal{d} \bm{y}_{R_k}\right|.
\end{equation*}

It is interesting to highlight that the evaluations of the junction tree of DEUNs and IDs follow the same backward inductive routine, formalized in Theorem \ref{theo2}, which sequentially absorbs a leaf of the tree. Given our definition of the cliques potentials, this absorption for DEUNs entails an updating of the utility potential only, which consist of a $\circ$ product. In contrast, for standard IDs this operation corresponds to a simple sum. To see this suppose that for a clique $C\in\mathcal{C}$, $\Psi_{C}=\sum_{j\in C\setminus S}k_ju(y_j)$ and $\Psi_{\mathcal{T}}=\sum_{C\in\mathcal{C}}\Psi_C$. The absorption of a clique $C_j$, supposing $C_j$ only includes chance nodes, into its parent $C_i$ in an ID with these potentials then changes $\Psi_{C_i}$ to
\begin{equation}
\Psi_{C_i}+\int_{\mathbb{Y}_{C_j}}\Psi_{C_j}\Phi_{C_j}\textnormal{d} \bm{y}_{C_j}.
\label{eq:absID}
\end{equation}
Equation (\ref{eq:absID}) can be seen to be almost identical to equation (\ref{eq:absorption}) which specifies the absorption step in DEUNs. The only difference lies in the different operation: a sum for IDs and a $\circ$ product for DEUNs.

\section{An application in food security}
\label{section8}
To provide an additional illustration of how the algorithms for the computation of expected utilities in DEUNs work in practice, we compute the expected utilities of a DEUN  applied to the field of food security. Food insecurity, defined as the \lq\lq{l}imited or uncertain availability of nutritionally adequate and safe foods or limited or uncertain ability to acquire acceptable foods in socially acceptable ways\rq\rq{} \cite{usda}, is not only an endemic issue in third world countries, but also a growing threat to wealthy nations. To support UK local governments to tackle the complexity of the evaluation of various policies to  ensure household food security, we have started building a probabilistic decision support tool modelling the food system.

\subsection{Network structure}
After a series of decision conferences with local authorities, stakeholders and potential decision makers, \citet{Barons2017} identified three areas that are impacted by increasing household food insecurity: educational attainment ($Y_E$), health ($Y_H$) and social cohesion ($Y_S$). Of course the cost ($Y_C$) associated to the enactment of any policy is deemed relevant in this domain. Measurable indexes were then developed for each of these areas - for instance, educational attainment is assessed by the percentage of pupils not failing a combination of UK school examinations. Suppose these indexes take values in $[0,100]$. Details about the form of the various attributes are beyond the scope of this paper and we refer to \citet{Barons2017} for a discussion of these.   

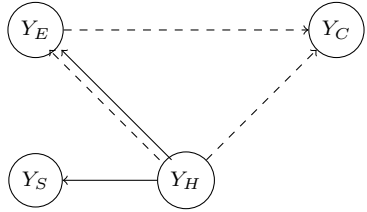
\begin{figure}
\begin{center}
\begin{tikzpicture}[node distance=4cm]
\node[draw,circle] (1) at (0,0) {$Y_S$};
\node[draw,circle](5) at (2,0) {$Y_H$};
\node[draw,circle] (3) at (0,2) {$Y_E$};
\node[draw,circle] (2) at (4,2) {$Y_C$};
\draw[->,sloped, anchor=center, above,near start] ([xshift=0.08cm]5.north west) to  node {} ([xshift=0.08cm]3.south east);
\draw[->,dashed,sloped, anchor=center, below,near start] ([xshift=-0.08cm]5.north west) to  node {} ([xshift=-0.08cm]3.south east);
\draw[->,sloped, anchor=center, above,near start] (5) to  node {} (1);
\draw[->,dashed,sloped, anchor=center, above,near start] (5) to  node {} (2);
\draw[->,dashed,sloped, anchor=center, above,near start] (3) to  node {} (2);
\end{tikzpicture}    
\end{center}
\caption{DEUN representing the food security example of Section \ref{section8}.\label{fig:food1}}
\end{figure}

Of course such a decision support system needs to model the probabilistic dependence over a much larger vector of variables that need to be accounted for in a reliable description of the food system. But for the illustrative purposes of this example, we assume the dependence structure between the four indexes of above is summarized by the DEUN in Figure \ref{fig:food1}. This states that the variable cost is independent of all others and that, given a specific value of the health index, educational attainment and social cohesion are independent. For the preferential part although a plausible assumption might be that the utilities of both health and social cohesion do not change when all the other attributes are varied,  the utility of various levels of educational attainment did appear to sometimes be a function of health. Similarly, the utility of the costs associated to policies' implementations appeared to be a function of both educational attainment and health. These assumptions are represented in the DEUN in Figure \ref{fig:food1} by the dashed arcs, depicting an underlying directional utility diagram.  

For this illustrative example we consider a decision space $\mathbb{D}$ including three policies: either an increase ($d_0$), a decrease $(d_1)$ or not a change $(d_2)$ of the number of pupils eligible for free school meals nationally. UK government has already implemented this type of policy to give pupils a healthy start in life, since evidence seems to point towards an improvement of development and social skills of young children  that eat a healthy meal together at lunchtime \citep{Kitchen2013}. In this setting, we define the variables $Y_E$, $Y_H$ and $Y_S$ as the variation in two years time of the corresponding current index value, whilst $Y_C$ is the change in the percentage of the government budget for the free school meal program.  We assume that each policy directly influences $Y_H$, $Y_E$ and $Y_C$, whilst $Y_s$ is only affected indirectly by a decision taken.

Initial discussions during the elicitation process suggested that a simple Normal regression model could be sufficient to depict the probabilistic part of the system. This is defined by the distributions
\begin{align*}
Y_H&\sim \mathcal{N}(\theta_{0H}^d,\sigma_H^d), \hspace{2cm} Y_E|Y_H\sim\mathcal{N}(\theta_{0E}^d+\theta_{HE}^dY_H,\sigma_E^d),\\
Y_C&\sim\mathcal{N}(\theta_{0C}^d,\sigma_C^d), \hspace{2cm} Y_S|Y_H\sim\mathcal{N}(\theta_{0S}+\theta_{HS}Y_H,\sigma_S),
\end{align*}
where the parameters $\theta$ and $\sigma$ take values in $\mathbb{R}$ and $\mathbb{R}_+$ respectively and a superscript $d$ denotes a different parameter value for each available policy. Notice that the above definitions are compatible with the underlying BN of Figure \ref{fig:food1}.

\begin{table}
\begin{center}
\begin{tabular}{|l|l|}
\hline
$\dot{u}(y_C|y_E^0,y_H^0)=\exp(-\delta_C^{00}y_C)$&$\dot{u}(y_E|y_H^0)=\exp(\delta_E^{0}y_E)$\Tstrut\Bstrut\\
$\dot{u}(y_C|y_E^0,y_H^*)=1-\exp(\delta_C^{0*}y_C)$&$\dot{u}(y_E|y_H^*)=\exp(\delta_E^{*}y_E)$\Tstrut\Bstrut\\
$\dot{u}(y_C|y_E^*,y_H^0)=1-\exp(\delta_C^{*0}y_C)$& $\dot{u}(y_H)=\exp(\delta_Hy_H)$\Tstrut\Bstrut\\
$\dot{u}(y_C|y_E^*,y_H^*)=1-\exp(\delta_C^{**}y_C)$& $\dot{u}(y_S)=\exp(\delta_Sy_S)$\Tstrut\Bstrut\\
\hline
\end{tabular}
\end{center}
\caption{Un-normalized utility functions of the free school meals example. \label{table:uti}}
\end{table}

\begin{figure}
\begin{center}
\includegraphics[scale=0.25]{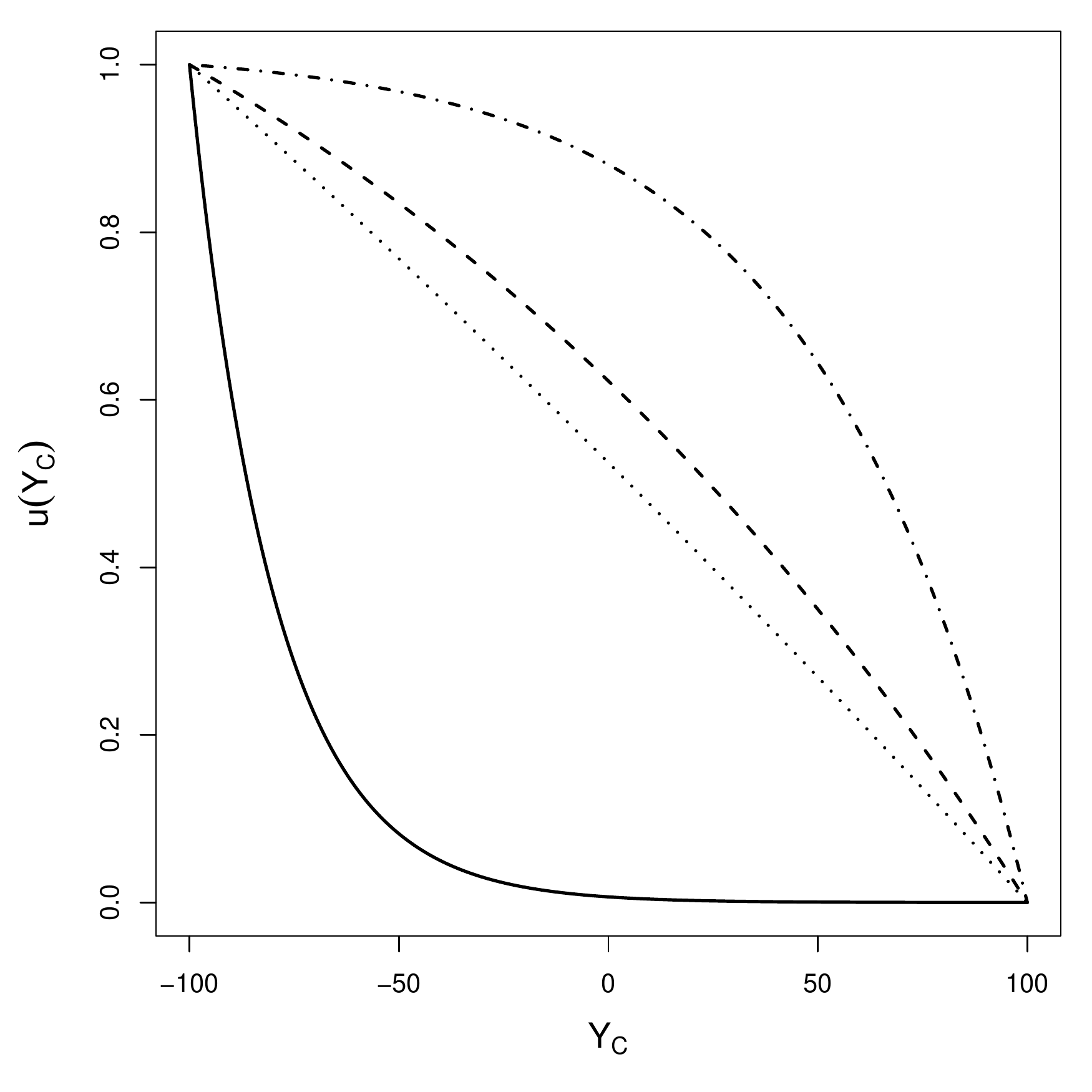}
\end{center}
\caption{Utility functions of $Y_C|\{Y_E,Y_H\}^{0*}$: full line - $y_E^0,y_H^0$; dotted line - $y_E^*,y_H^0$; dashed line - $y_E^0,y_H^*$; dotted/dashed line - $y_E^*,y_H^*$; \label{fig:yc}}
\end{figure}

We assume the utilities to be exponentials and of the form specified in Table \ref{table:uti}, where the parameters $\delta$ take values $\mathbb{R}_+$. These then need to be normalized. For an attribute $Y$ this can be done using the formula $u(y)=(\dot{u}(y)-m)/(M-m),$ where $\dot{u}$ is the un-normalized utility function, $m=\min(\dot{u}(y))$ and $M=\max (\dot{u}(y))$ . So for example Figure \ref{fig:yc} shows the normalized version of the utility functions of costs conditional on the boundary values of educational attainment and health, for a specific choice of the parameters $\delta$. Again these utility definitions are compatible with the DEUN structure of Figure \ref{fig:food1}. 

\subsection{The algorithm}
Given the definitions of the DEUN structure and of the specific form of the probability and utility functions, we can now proceed with an illustration of our evaluation algorithm. Since the DEUN in Figure \ref{fig:food1} is non-decomposable, for its evaluation we need to use the algorithm in Theorem \ref{theo1}. There are many variable orderings that the algorithm could follow, but we here choose the sequence $(Y_E,Y_S,Y_H,Y_C)$.

First notice that the vector $\bm{u}_E$ consists of the four entries $u(y_E|y_H^0)$, $\hat{u}(y_E|y_H^0)$, $u(y_E|y_H^*)$ and $\hat{u}(y_E|y_H^*)$. The first step of the algorithm, as formalized in equation (\ref{eq:theo1}), computes the expectation of these utilities with respect to the conditional probability function of $Y_E$ given $Y_H$. This consists of the computation of the moment generating function of a normal random variable. Recall that for a normal random variable $Y$ with mean $\mu$ and variance $\sigma$ and a $t\in\mathbb{R}$ we have that $\mathbb{E}(\exp(tY))=\exp(t\mu+0.5t^2\sigma^2)$. Thus  
\[
\overline{\bm{u}}_E=\left(
\frac{E_d^0-m_E^0}{M_E^0-m_E^0},\,\,\frac{M_E^0-E_d^0}{M_E^0-m_E^0},\,\,
 \frac{E_d^*-m_E^*}{M_E^*-m_E^*},\,\, \frac{M_E^*-E_d^*}{M_E^*-m_E^*}  
 \right)
\]
where $M_E$ and $m_E$, with the appropriate superscript, denote the maximum and the minimum of the utility function respectively, and
\begin{align*}
&E_d^0=\exp(\delta_E^0\theta_{0E}^d+\delta_{E}^0\theta_{HE}^dY_H+0.5(\delta_E^0\sigma_E^d)^2),\\ 
&E_d^*=\exp(\delta_E^*\theta_{0E}^d+\delta_{E}^*\theta_{HE}^dY_H+0.5(\delta_E^*\sigma_E^d)^2).
\end{align*}

Next the algorithm considers the node $Y_S$.  As specified by equation (\ref{eq:theo11}), it first computes $\overline{\bm{u}}_E\circ \bm{u}_S$, where $\bm{u}_S=(u(y_S),1-u(y_S))$. This $\circ$ product is given by $(\overline{\bm{u}}_Eu(y_S),\overline{\bm{u}}_E(1-u(y_S)))$ since $\overline{\bm{u}}_E$ is not a function of $Y_S$. Then equation (\ref{eq:theo11}) computes $\overline{\bm{u}}_S$ as the expectation of each entry of $\overline{\bm{u}}_E\circ \bm{u}_S$ with respect to  $p(y_S|y_H)$.  This gives the vector 
\[\overline{\bm{u}}_S=\left(\overline{\bm{u}}_E(S-m_S)/(M_S-m_S),\overline{\bm{u}}_E(M_S-S)/(M_S-m_S)\right),\]
 where $S=\exp(\delta_S(\theta_{0S}+\theta_{HS}y_H)+0.5\delta_S^2\sigma_S^2)$.

At this point the algorithm moves to $Y_H$ and computes $\overline{\bm{u}}_S\circ \bm{u}_H$. Notice that $\overline{\bm{u}}_S$ is already a function of $Y_H$. Specifically the first, second, fifth and sixth entries of $\overline{\bm{u}}_S$ refer to $y_H^0$ and therefore need to be multiplied by $1-u(y_H)$, whilst the others need to be multiplied by $u(y_H)$. Then equation (\ref{eq:theo11}) computes the expectation of this product with respect to $p(y_H)$ giving an 8-dimensional vector $\overline{\bm{u}}_H$ whose entries $\overline{\bm{u}}_H(i)$, $i\in[8]$,
are given in Table \ref{table:boh} with indeterminates defined in Table \ref{table:uff}.

\begin{table}
\begin{center}
\scalebox{0.7}{
\begin{tabular}{|l|}
\hline
$\overline{\bm{u}}_H(1)=(M_H\overline{ES}_d^0-\overline{ESH}_d^0+m_S(\overline{EH}_d^0-M_H\overline{E}_d^0)+m_E^0(\overline{SH}_d-M_H\overline{S}+m_S(M_H-\overline{H}_d)))/((M_E^0-m_E^0)(M_S-m_S)(M_H-m_H))$\Tstrut\Bstrut\\
$\overline{\bm{u}}_H(2)=(M_E^0(M_H\overline{S}-\overline{SH}_d+m_S(\overline{H}_d-M_H))-M_H\overline{ES}_d^0+\overline{ESH}_d^0+m_S(M_H\overline{E}_d^0-\overline{EH}_d^0))/((M_E^0-m_E^0)(M_S-m_S)(M_H-m_H))$\Tstrut\Bstrut\\
$\overline{\bm{u}}_H(3)=(\overline{ESH}_d^*-m_H\overline{ES}_d^*+m_S(m_H\overline{E}_d^*-\overline{EH}_d^*)+m_E^*(m_S\overline{H}_d-\overline{SH}_d^*+m_H(\overline{S}-m_S)))/((M_E^*-m_E^*)(M_S-m_S)(M_H-m_H))$\Tstrut\Bstrut\\
$\overline{\bm{u}}_H(4)=(M_E^*(\overline{SH}_d-m_H\overline{S}+m_S(m_H-\overline{H}_d))-\overline{ESH}_d^*+m_H\overline{ES}_d^*+m_S(\overline{EH}_d^*-m_H\overline{E}_d^*))/((M_E^*-m_E^*)(M_S-m_S)(M_H-m_H))$\Tstrut\Bstrut\\
$\overline{\bm{u}}_H(5)=(M_H(M_S\overline{E}_d^0-\overline{ES}_d^0)+\overline{ESH}_d^0-M_S\overline{EH}_d^0+m_E^0(M_S(\overline{H}_d-M_H)+M_H\overline{S}-\overline{SH}_d))/((M_E^0-m_E^0)(M_S-m_S)(M_H-m_H))$\Tstrut\Bstrut\\
$\overline{\bm{u}}_H(6)=(M_E^0(M_S(M_H-\overline{H}_d)-M_H\overline{S}+\overline{SH}_d)+M_S(\overline{EH}_d^0-M_H\overline{E}_d^0)+M_H\overline{ES}_d^0-\overline{ESH}_d^0)/((M_E^0-m_E^0)(M_S-m_S)(M_H-m_H))$\Tstrut\Bstrut\\
$\overline{\bm{u}}_H(7)=(M_S(\overline{EH}_d^*-m_H\overline{E}_d^*)-\overline{ESH}_d^*+m_H\overline{ES}_d^*+m_E^*(\overline{SH}-M_S\overline{H}+m_H(M_S-\overline{S})))/((M_E^*-m_E^*)(M_S-m_S)(M_H-m_H))$\Tstrut\Bstrut\\
$\overline{\bm{u}}_H(8)=(M_E^*(M_S(\overline{H}_d-m_H)-\overline{SH}_d+m_H\overline{S})+\overline{ESH}_d^*-m_H\overline{ES}_d^*+M_S(m_H\overline{E}_d^*-\overline{EH}_d^*))/((M_E^*-m_E^*)(M_S-m_S)(M_H-m_H))$\Tstrut\Bstrut\\
\hline
\end{tabular}}
\end{center}
\caption{Entries of $\bar{\bm{u}}_H$. \label{table:boh}}
\end{table}

\begin{table}
\begin{center}
\scalebox{0.75}{
\begin{tabular}{|l|}
\hline
$\overline{S}=\exp(\delta_S(\theta_{0S}+\theta_{HS}\theta_{0H}^d)+0.5\delta_S^2(\sigma_S^2+(\theta_{HS}\sigma_H^d)^2))$\Tstrut\Bstrut\\
$\overline{E}_d^0=\exp(\delta_E^0(\theta_{0E}^d+\theta_{HE}^d\theta_{0H}^d)+0.5(\delta_E^0)^2((\sigma_E^d)^2+(\theta_{HE}^d\sigma_H^d)^2))$\Tstrut\Bstrut\\
$\overline{H}_d=\exp(\delta_H\theta_{0H}^d+0.5(\delta_H\sigma_H^d)^2)$\Tstrut\Bstrut\\
$\overline{SH}_d=\exp(\delta_S\theta_{0S}+0.5\delta_S^2\sigma_S^2+(\delta_S\theta_{HS}+\delta_H)\theta_{0H}^d+0.5(\delta_S\theta_{HS}+\delta_H)^2(\sigma_H^d)^2)$\Tstrut\Bstrut\\
$\overline{EH}_d^0=\exp(\delta_E^0\theta_{0E}^d+0.5(\delta_E^0\sigma_E^d)^2+(\delta_E^0\theta_{HE}^d+\delta_H)\theta_{0H}^d+0.5(\delta_E^0\theta_{HE}^d+\delta_H)^2(\sigma_H^d)^2)$\Tstrut\Bstrut\\
$\overline{ES}_d^0=\exp(\delta_S\theta_{0S}+\delta_{E}^0\theta_{0E}^d+0.5(\delta_S^2\sigma_S^2+(\delta_E^0\sigma_E^d)^2)+(\delta_S\theta_{HS}+\delta_E^0\theta_{HE}^d)\theta_{0H}^d+0.5(\delta_S\theta_{HS}+\delta_E^0\theta_{HE}^d)^2(\sigma_H^d)^2)$\Tstrut\Bstrut\\
$\overline{ESH}_d^0=\exp(\delta_S\theta_{0S}+\delta_{E}^0\theta_{0E}^d+0.5(\delta_S^2\sigma_S^2+(\delta_E^d\sigma_E^d)^2)+(\delta_S\theta_{HS}+\delta_E^0\theta_{HE}^d+\delta_H)\theta_{0H}^d+0.5(\delta_S\theta_{HS}+\delta_E^0\theta_{HE}^d+\delta_H)^2(\sigma_H^d)^2)$\Tstrut\Bstrut\\
\hline
\end{tabular}}
\end{center}
\caption{Definition of the indeterminates in Table \ref{table:boh}, where $\overline{E}_d^*$, $\overline{EH}_d^*$, $\overline{ES}_d^*$ and $\overline{ESH}_d^*$ are similarly defined. \label{table:uff}}
\end{table}

The algorithm then moves to node $Y_C$. Since $\overline{\bm{u}}_H$ is not a function of $Y_C$, $\overline{\bm{u}}_H\circ\bm{u}(y_C|y_E^{0*},y_H^{0*})$ returns the elements
\[
\begin{array}{ccccc}
\overline{\bm{u}}_H(1)u(y_C|y_E^*,y_H^0),&&&&\overline{\bm{u}}_H(2)u(y_C|y_E^0,y_H^0),\\
\overline{\bm{u}}_H(3)u(y_C|y_E^*,y_H^*),&&&&\overline{\bm{u}}_H(4)u(y_C|y_E^0,y_H^*),\\
\overline{\bm{u}}_H(5)u(y_C|y_E^*,y_H^0),&&&&\overline{\bm{u}}_H(6)u(y_C|y_E^0,y_H^0),\\
\overline{\bm{u}}_H(7)u(y_C|y_E^*,y_H^*),&&&&\overline{\bm{u}}_H(8)u(y_C|y_E^0,y_H^*),\\
\overline{\bm{u}}_H(1)\hat{u}(y_C|y_E^*,y_H^0),&&&&\overline{\bm{u}}_H(2)\hat{u}(y_C|y_E^0,y_H^0),\\
\overline{\bm{u}}_H(3)\hat{u}(y_C|y_E^*,y_H^*),&&&&\overline{\bm{u}}_H(4)\hat{u}(y_C|y_E^0,y_H^*),\\
\overline{\bm{u}}_H(5)\hat{u}(y_C|y_E^*,y_H^0),&&&&\overline{\bm{u}}_H(6)\hat{u}(y_C|y_E^0,y_H^0),\\\overline{\bm{u}}_H(7)\hat{u}(y_C|y_E^*,y_H^*),&&&&\overline{\bm{u}}_H(8)\hat{u}(y_C|y_E^0,y_H^*).
\end{array}
\]
The expectation of the above terms with respect to $p(y_C)$ then follows by simply applying the moment generating function relationships for normal random variables, since $\overline{\bm{u}}_H$ is not a function of $Y_C$. We denote the resulting vector as $\overline{\bm{u}}_C=(\overline{\bm{u}}_C(i))_{i\in[16]}$.

As formalized in equation (\ref{eq:theo12}), the algorithm then terminates by taking the sum of the element of $\overline{\bm{u}}_C$ multiplied by the appropriate weighting term $u(\bm{y}^{0*})$. Specifically, the overall expected utility for a decision $d\in\mathbb{D}$ equals the sum of the terms

\[
\begin{array}{lllll}
\overline{\bm{u}}_C(1)u(y_E^*,y_S^*,y_H^0,y_C^*),&&&&\overline{\bm{u}}_C(2)u(y_E^0,y_S^*,y_H^0,y_C^*),\\\overline{\bm{u}}_C(3)u(y_E^*,y_S^*,y_H^*,y_C^*),&&&&\overline{\bm{u}}_C(4)u(y_E^0,y_S^*,y_H^*,y_C^*),\\
\overline{\bm{u}}_C(5)u(y_E^*,y_S^0,y_H^0,y_C^*),&&&&\overline{\bm{u}}_C(6)u(y_E^0,y_S^0,y_H^0,y_C^*),\\\overline{\bm{u}}_C(7)u(y_E^*,y_S^0,y_H^*,y_C^*),&&&&\overline{\bm{u}}_C(8)u(y_E^0,y_S^0,y_H^*,y_C^*),\\
\overline{\bm{u}}_C(9)u(y_E^*,y_S^*,y_H^0,y_C^0),&&&&\overline{\bm{u}}_C(10)u(y_E^0,y_S^*,y_H^0,y_C^0),\\\overline{\bm{u}}_C(11)u(y_E^*,y_S^*,y_H^*,y_C^0),&&&&\overline{\bm{u}}_C(12)u(y_E^0,y_S^*,y_H^*,y_C^0),\\
\overline{\bm{u}}_C(13)u(y_E^*,y_S^0,y_H^0,y_C^0),&&&&\overline{\bm{u}}_C(14)u(y_E^0,y_S^0,y_H^0,y_C^0),\\\overline{\bm{u}}_C(15)u(y_E^*,y_S^0,y_H^*,y_C^0),&&&&\overline{\bm{u}}_C(16)u(y_E^0,y_S^0,y_H^*,y_C^0).\\
\end{array}
\]

Notice that the overall expected utility is a highly non-linear function of the problem's parameters. But it has a closed-form expression and this form is the same for all available decisions. Thus the identification of an optimal strategy can then be carried out by simply plugging-in the different numerical specifications associated to different policies. In Appendix \ref{app} we give plausible values to the parameters of the free school meal example. For such values, the decision $d_0$ of increasing the number of eligible pupils would be optimal having expected utility score $0.29$, compared to $0.19$ and $0.21$ for policies $d_1$ and $d_2$ respectively.  

\section{Discussion}
\label{section9}
Graphical representations of both probabilistic and preferential independences have received great attention in the literature. However, so far very little effort has been applied to the study of how probabilistic and preferential graphical models could be combined to provide a graphical representation of the expected utility structure of a decision problem. In this paper we presented one of the first attempts to formally define a network model depicting both the probabilistic and the utility relationships for a random vector of attributes. We have demonstrated here how such a graphical representation then provides a framework for the fast computation of the overall expected utility through a variable elimination algorithm over the junction tree of a DEUN. 

Whilst the constraint of having only directed probabilistic edges is very often met in practice, and indeed BNs are the most common probabilistic graphical model, restricting the class of underlying utility diagrams to only directional ones may be unreasonable in some applications. Intuitively, a more general utility factorization without the constraint of a directional utility diagram can lead to a distributed computation of expected utilities by coupling generic utility diagrams with probabilistic chain graphs \citep{Lauritzen1996}. Propagation algorithms also exist for this model class and therefore adaptations of these could enable the computation of expected utilities in this more general class of models. 

Lastly, DEUNs could also be generalized to include decision nodes and therefore fully represent the structure of a DM's decision problem, just as influence diagrams extend BN models. We envisage that the evaluation of such a network could be performed by algorithms that share many features with the ones presented here, but that are also equipped with optimization steps over decision spaces.

\section*{Acknowledgements}
The work of M. Leonelli was supported by Capes, whilst J.Q. Smith was partly supported by EPSRC grant EP/K039628/1.

\bibliographystyle{imsart-nameyear}
\bibliography{bib}
\begin{appendix}
\section{Numerical specifications for the food security example}

\label{app}
\vspace{0.5cm}

\begin{center}
\begin{tabular}{c|c|c|c|c|c|c|c}
&$\theta_{0H}^d$& $\sigma_H^d$&$\theta_{0C}$&$\sigma_C^d$&$\theta_{0E}^d$& $\sigma_E^d$&$\theta_{HE}^d$\\
\hline
$d_0$&1.5&5&30&8&5&40&7\Tstrut\Bstrut\\
$d_1$&-2&4&-5&5&-6&20&2\Tstrut\Bstrut\\
$d_2$&-0.5&3&10&4&3&15&7\Tstrut\Bstrut
\end{tabular}

\vspace{0.5cm}

\begin{tabular}{|ccc|}
\hline
$\delta_C^{00}=0.05$,&$\delta_E^0=0.01$,&$\theta_{0S}=5$,\Tstrut\Bstrut\\
$\delta_C^{0*}=0.005$,&$\delta_E^*=0.005$,&$\theta_{HS}=17$,\Tstrut\Bstrut\\
$\delta_C^{*0}=0.001$,&$\delta_S=0.01$,&$\sigma_{S}=20$,\Tstrut\Bstrut\\
$\delta_C^{**}=0.02$,&$\delta_H=0.02$,&\Tstrut\Bstrut\\
\hline
\end{tabular}
\end{center}

\vspace{0.5cm}

\begin{center}
\begin{tabular}{|cc|}
\hline
$u(y_E^0,y_S^0,y_H^0,y_C^0)=0$&$u(y_E^*,y_S^0,y_H^0,y_C^0)=0.25$\Tstrut\Bstrut\\
$u(y_E^0,y_S^*,y_H^0,y_C^0)=0.2$&$u(y_E^*,y_S^*,y_H^0,y_C^0)=0.5$\Tstrut\Bstrut\\
$u(y_E^0,y_S^0,y_H^*,y_C^0)=0.5$&$u(y_E^*,y_S^0,y_H^*,y_C^0)=0.75$\Tstrut\Bstrut\\
$u(y_E^0,y_S^*,y_H^*,y_C^0)=0.7$&$u(y_E^*,y_S^*,y_H^*,y_C^0)=0.85$\Tstrut\Bstrut\\
$u(y_E^0,y_S^0,y_H^0,y_C^1)=0.05$&$u(y_E^*,y_S^0,y_H^0,y_C^1)=0.3$\Tstrut\Bstrut\\
$u(y_E^0,y_S^*,y_H^0,y_C^1)=0.25$&$u(y_E^*,y_S^*,y_H^0,y_C^1)=0.55$\Tstrut\Bstrut\\
$u(y_E^0,y_S^0,y_H^*,y_C^1)=0.55$&$u(y_E^*,y_S^0,y_H^*,y_C^1)=0.8$\Tstrut\Bstrut\\
$u(y_E^0,y_S^*,y_H^*,y_C^1)=0.75$&$u(y_E^*,y_S^*,y_H^*,y_C^1)=1$\Tstrut\Bstrut\\
\hline
\end{tabular}
\end{center}
\end{appendix}
\end{document}